\documentclass[10pt,amssymb,twocolumn,notitlepage]{article}

\usepackage{epigraph}
\usepackage[affil-it]{authblk}
\usepackage{dcolumn}
\usepackage[stable]{footmisc}
\usepackage{blkarray}
\usepackage{bm}
\usepackage[mathscr]{euscript}
\usepackage[font=scriptsize]{caption}
\usepackage{subcaption,graphicx}
\usepackage[table]{xcolor}
\usepackage[margin=1.7cm]{geometry}
\usepackage{tcolorbox}
\usepackage{hyperref}
\usepackage{gensymb}
\usepackage[font=footnotesize]{caption}
\usepackage{hyperref}
\hypersetup{
    colorlinks=true,
    linkcolor=blue,
    filecolor=magenta,      
    urlcolor=blue,
}
\usepackage{steinmetz}
\usepackage{amssymb}
\usepackage{makecell}
\usepackage{algorithm,lipsum,xcolor,caption}
\usepackage{enumitem}
\usepackage{amsmath,esint}
\usepackage{fancyvrb}
\usepackage{xcolor}
\usepackage{arydshln}
\usepackage{mathtools}

\definecolor{newblue}{rgb}{0.0, 0.28, 0.67}
\definecolor{newgreen}{rgb}{0.13, 0.55, 0.13}
\definecolor{newred}{rgb}{0.87, 0.72, 0.53}

\newcommand{\R}{\mathbb{R}}
\newcommand{\lbar}{\{\kern-0.5ex|}
\newcommand{\rbar}{|\kern-0.5ex\}}
\usepackage{stackengine}

\definecolor{newblue}{rgb}{0.0, 0.28, 0.67}
\definecolor{newgreen}{rgb}{0.13, 0.55, 0.13}
\definecolor{newred}{rgb}{0.87, 0.72, 0.53}

\title{On Similarity}
\author{Luciano da Fontoura Costa \\ \emph{luciano@ifsc.usp.br}}
\affil{S\~ao Carlos Institute of Physics -- DFCM/USP} 
\date{25rd Oct 2021}

\begin{document}

\twocolumn[
\begin{@twocolumnfalse}
    \maketitle
    \begin{abstract}
    The objective quantification of similarity between two mathematical structures
    constitutes a recurrent issue in science and technology.  In the present work,
    we developed a principled approach that took the Kronecker's delta function
    of two scalar values as the prototypical reference for similarity quantification and then derived
    for more yielding indices, three of which bound between 0 and 1.   Generalizations
    of these indices to take into account the sign of the scalar values were then presented
    and developed to multisets, vectors, and functions in real spaces.  Several important
    results have been obtained, including the interpretation  of the Jaccard index as 
    a yielding implementation of the Kronecker's delta function.  When generalized to
    real functions, the four described similarity indices become respective functionals,
    which can then be employed to obtain associated operations of convolution and
    correlation.
    \end{abstract}
\end{@twocolumnfalse} \bigskip
]

\setlength{\epigraphwidth}{.49\textwidth}
\epigraph{`Springtime, always plentiful of most diverse similarities.'}
{\emph{LdaFC}}

\section{Introduction}

It is often mentioned that one of the most important aspects of science is the quantification of the
physical world structures and phenomena, through respective measurements, so as to 
allow the development of objective theories.   While this is certainly true, there is a complementary
aspect to taking measurements, and this concerns \emph{comparing} and \emph{ordering}
the obtained quantifications so as to be able to take decisions on the most plausible models and
explanations.  

For instance, given a model, its ability to account for the respectively modeled
systems consists of comparing not only scalar values, but vectors, matrices, functions, as well
as potentially any other mathematica structure.  Indeed, the own validation of models
and theories rely critically on several logical and quantitative indications of similarity.

In addition to the critically important role of comparisons in science, living beings also continuously
rely on comparing entities, be then a received stimuli or more complex mental representations.
For instance, we humans are always comparing today's weather with those of other times.  

Comparing things in an objective quantitative manner involves the adoption of one or more
measurements of similarity or distance between pairs of values, of which the Euclidean distance
seems to have a particular importance.   

On subsequent scales, measurements are also required
for comparing sets of objects, and here the cosine similarity and Jaccard indices (e.g.~\cite{CostaJaccard,CostaMset}) are often
employed.  At an even higher level, we need approaches capable of quantifying the similarity
between functions, in which case the inner product, and the respectively derived operations of
convolution  and correlation, are often adopted (e.g.~\cite{brigham:1988,DudaHart,Koutrombas,shapebook}).

Given that similarity and distance are intrinsically interrelated, including the fact that one can
often be derived from the other, the present work will focus only on similarity measurements, 
with the obtained results being immediately extensible to distances.

Interestingly, the Euclidian distance, cosine similarity and inner product all share a same aspect,
which consists in being based on products between pairs of values.   As such, these approaches
can be said to have a \emph{second order} nature ($x.x = x^2$).   
However, there is a virtually infinite number of other possible distance and similarity measurements,
including those based on minimum, maximum and absolute values.

In the present work, we aim at developing a principled approach in which we start 
by contemplating the similarity between two scalar real values $x$ and $y$, from
which it is concluded that the Kronecker's delta function provides a prototypical 
reference.  However, given that this approach is too strict, it becomes necessary to
relax the Kronecker's delta function criterion so as to obtain more yielding respective
similarity indices.  Four main possibilities are identified, three of which being
suitably bound in the interval $[0,1]$, being denominated $s_1$, $s_2$, $s_3$,
and $s_4$.

Then, by using concepts derived from~\cite{CostaJaccard,CostaMset}, we describe how these 
four indices can be generalized in order to provide additional information about the
relative alignment between the two compared scalar values, yielding 4 respective versions
of the adopted indices.

These indices are then further generalized, again by considering the results in~\cite{CostaJaccard,CostaMset}, to multisets (e.g.~\cite{Hein,Knuth,Blizard,Blizard2,Thangavelu,Singh}), vectors, and real
functions.  Though respective extensions to other mathematical structures including
matrices, graphs, and scalar and vector fields are analogous, we do not develop
these possibilities in the present work.

Two other similarity indices, namely the interiority (or homogeneity) and coincidence
indices~\cite{CostaJaccard,CostaMset,CostaComparing}, are also presented in their
generalized versions for functions.  In particular, the coincidence index has been
found to present enhanced performance in important tasks such as pattern 
recognition~\cite{CostaComparing} as well as when extended to act as quantifiers
of joint variation between random variables~\cite{CostaJaccard,CostaCCA}.

When generalized to real function spaces, the four indices become functionals and,
as such, can be combined in several manners and also used to implement respective
convolution and correlation binary operations between functions.  

Among the several interesting results obtained, we have that the indices proposed
in~\cite{CostaJaccard}, especially the coincidence and addition-based mset
Jaccard indices, actually corresponds to the generalizations of the scalar indices
$s_1$ and $s_2$.

We start by deriving the four similarity index from the Kronecker's delta function,
and proceed by presenting their extension to negative values and further generalization
to multisets, vectors, and functions.   Generalizations of the interiority and continuity
indices to multisets, vectors and functions possibly taking negative values are then presented,
which is followed by the presentation of the employment of all considered indices to
define respective convolutions and correlations.

\section{Pairwise Similarities in $\R$}

Before proceeding in depth with any current study of distances, it is important to 
state as objectively as possible what is being meant by similarity.  Unlike the
concept of distance, which is ubiquitously associated to the concept of
Euclidean distance, there seems to be less consensus regarding what similarity
means.  

\begin{figure}[h!]  
\begin{center}
   \includegraphics[width=0.5\linewidth]{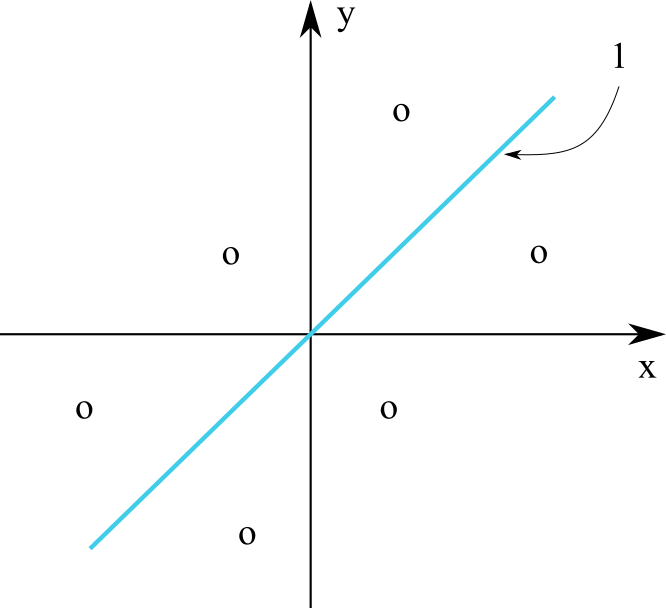}  
    \caption{The most strict quantification of the similarity between two real values $x$ and $y$, 
    implemented via a similarity binary operator $\delta_{x,y}$ that corresponds to the
    Kronecker delta.   A non-zero result is obtained only in case $x=y$.}
    \label{fig:identity}
    \end{center}
\end{figure}

In this work, we will understand similarity between two values $x$ and $y$ in the sense of 
\emph{identity} between them.  Figure~\ref{fig:identity} illustrates the most strict approach
to quantifying the similarity between any two real values $x$ and $y$, which assigns 
1 whenever $x=y$, and $0$ otherwise.

\vspace{0.5cm}

For simplicity's sake, the case $(x=0,y=0)$ is not considered in this work.  Therefore,
additional checking and handling is required in situations in which these values are viable.

Mathematically, this strict similarity quantification corresponds to a continuous Dirac 
delta comb function $\delta_{x,y}$.

The problem with this approach evidently is that it is way too strict, so that it becomes
necessary to provide means for implementing some tolerance in the quantification.

The distance between any vector $\left[x,y\right]$ and the line $y =x$ can be
readily expressed as:
\begin{eqnarray}
   d\left(\vec{p}= \left[ x,y\right],\hat{u}=\left[\frac{1}{\sqrt{2}},\frac{1}{\sqrt{2}} \right] \right) 
    = \frac{|x-y|}{\sqrt{2}}
\end{eqnarray}

Observe that this function is not upper bound, i.e.~all we can say is 
that $0 \leq d\left( \vec{p}, \hat{u} \right) $.

A possible manner to bound this distance is by making:
\begin{equation}
    \tilde{d}\left( \vec{p}, \hat{u} \right)  = \frac{|x-y|}{\max \left\{ |x|, |y| \right\}}
\end{equation}

which now ensures that $0 \leq \tilde{d}\left( \vec{p}, \hat{u} \right) \leq1$.

Having a distance measurement normalized in the interval $[0,1]$ is of critical
importance because it allows us to derive a respective similarity distance 
simply as:
\begin{equation}
   s \left( \vec{p}, \hat{u} \right)  = 1-  \frac{|x-y|}{\max \left\{ |x|, |y| \right\}}
\end{equation}

It can be verified that:
\begin{equation}
   1-  \frac{|x-y|}{\max \left\{ |x|, |y| \right\}} =  \frac{2 \min\left\{ |x|,  |y| \right\} }{|x| + |y| } 
\end{equation}

which is a slightly  more convenient manner to express this similarity, which will
constitute one of the similarity index addressed in the present work:
\begin{equation}  \label{eq:s1}
  s_1(x,y) =  \frac{2 \min\left\{ |x|,  |y| \right\} }{|x| + |y| } 
\end{equation}

with $0 \leq s_1(x,y) \leq 1$.     

\begin{figure}[h!]  
\begin{center}
   \includegraphics[width=0.7\linewidth]{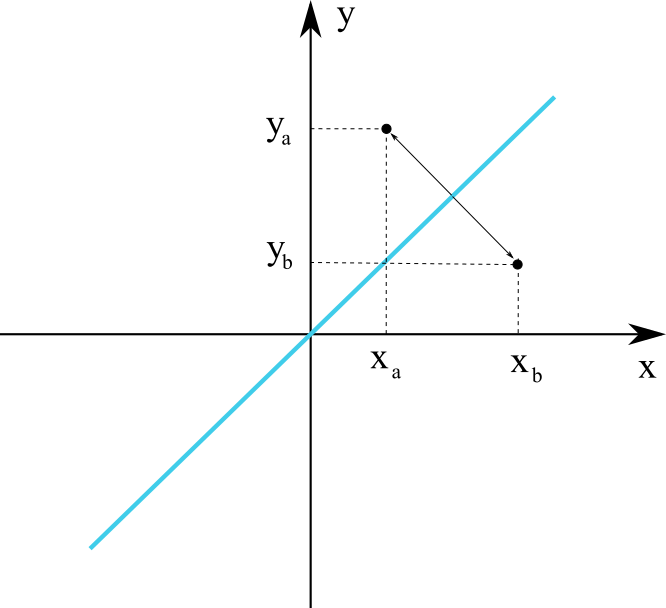}  
    \caption{Two pairs of values $(x_a,y_a)$ and $(x_b,y_b)$ are to be
    compared in terms of similarity. In the former case, we have that
    $\max\left\{ x_a,y_a \right\} = x_a$ and $\min\left\{ x_a,y_a \right\} = y_a$,
    with $x_a \neq y_a$, yielding a respective similarity smaller than one.
    In the other case, we will also have a similarity smaller than one for
    similar reasons.  Actually, because of the intrinsic symmetry in this case,
    the two obtained values of $s_1$ will be identical. }
    \label{fig:s1}
    \end{center}
\end{figure}

This similarity measurement is illustrated in Figure~\ref{fig:s1}.

\vspace{0.5cm}

The average between $|x|$ and $|y|$ can now be replaced by $\max \left\{ |x|, ||y|\right\}$,
yielding another normalized similarity index:
\begin{equation}
  s_2(x,y) = \frac{\min\left\{|x|, |y| \right\}}  {\max \left\{ |x|, |y| \right\} }
\end{equation}

with $0 \leq s_2(x,y) \leq 1$.

Yet another possible modification of the similarity index in Equation~\ref{eq:s1}
can be obtained by considering the product of functions:
\begin{equation}
     s_3 = \frac{ |x| |y| }{\left( \max \left\{ |x|,  |y| \right\} \right)^2 }   \\
\end{equation}

with $0 \leq s_3(x,y) \leq 1$.   

It is also interesting to consider the following unbound version of the index $s_3$:
\begin{equation}
    s_4 = |x|  |y|
\end{equation}

with $0 \leq s_4 \leq \infty$.

Though other similarity indices can be derived in analogous manner, the present
work will focus on the three indices $s_1$, $s_2$ and $s_3$ above.

Now, it is interesting to realize that the above indices loose information about the
relative signs of the involved quantities.  While this feature is suitable, and even
desired in some circumstances, it is important to have generalizations of the
three similarity indices derived above that can take into account the signs of
the involved quantities.

\begin{figure}[h!]  
\begin{center}
   \includegraphics[width=0.9\linewidth]{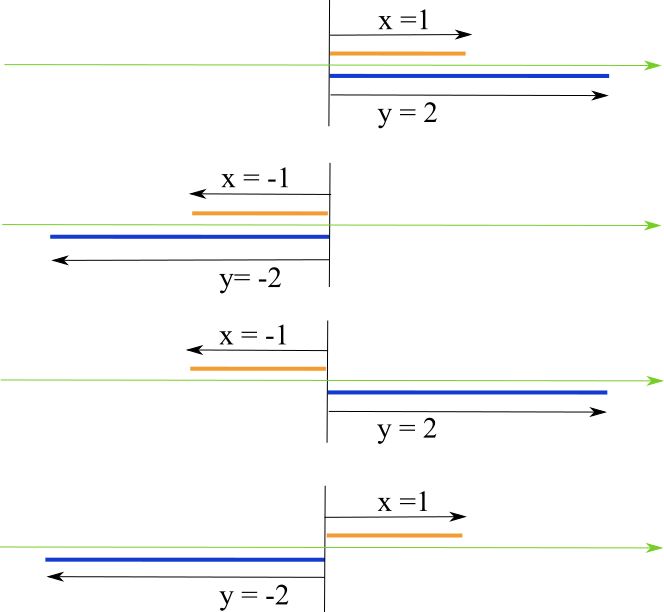}  
    \caption{The four main situations met when comparing two positions
    $x$ and $y$ along the real line $\R$.  It is often interesting to take
    into account whether the positions point toward the same or opposite directions. }
    \label{fig:neg}
    \end{center}
\end{figure}

Consider the situations depicted in Figure~\ref{fig:neg}.  Here, we have the four
situations which needed to be taken into account while generalizing the three
adopted similarity indices to cope with negative values.

\vspace{0.5cm}

The similarity sign should express whether the two positions point toward the same
direction, which case a positive similarity could be expected, or it they oppose
one another, yielding a respective negative similarity sign.  

The key to obtaining signed similarity consists in employing the following functions:
\begin{align}
  &s_{x} = \emph{sign}(x) \\
  &s_{y} = \emph{sign}(y) \\
  &s_{xy} = \emph{sign}(x) \; \emph{sign}(y)
\end{align}

We shall refer to the function $s_{xy}$ as the \emph{conjoint sign function}.

We can now generalize the three adopted similarity index to reflect the sign of the
values $x$ and $y$ as:
\begin{align}
     &s_1 = s_{xy} \frac{\min\left\{s_x x,  s_y y \right\} }{\max\left\{ s_x x, s_y y \right\} }   \\
     &s_2 = s_{xy} \frac{2 \min\left\{ s_x x,  s_y y \right\} }{s_x x + s_y y }    \\
     &s_3 =  \frac{  x  y }{\left( \max \left\{ s_x  x,   s_y y \right\} \right)^2 }    \\
     &s_4 = s_{xy} (s_x x s_y y) = s_{xy}^2 x y = xy
\end{align}

with $-1 \leq s_1, s_2, s_3 \leq 1$ and $-\infty \leq s_4 \leq \infty$.

For simplicity's sake, both the modulus and signed versions
of the three similarities will be henceforth referred to simply 
as $s_1$, $s_2$, and $s_3$, as the context
shall be enough to indicate how they are being applied.

In the context of polynomials, the product of two values $x$ and $y$
represents a second degree operation.  This operation has an
intrinsic characteristic in which the product of two numbers larger
than one tend to increase steeply with the magnitude of the values.
However, when two values with magnitude smaller than 1 are multiplied,
the resulting value is typically substantially reduced.
This characteristic is a direct consequence of the \emph{non-bilinearity} of
the product operation.

Interestingly, the results from $s_1$ to $s_4$ can be understood as
providing successively blurred versions of the Kronecker's delta
reference similarity functions.  Therefore, more strict quantifications
of similarity between two values $x$ and $y$ will be provided by
$s_1$ and $s_2$, while $s_3$ and $s_4$ represent particularly 
yielding alternatives. 

\begin{figure}[h!]  
\begin{center}
   \includegraphics[width=1\linewidth]{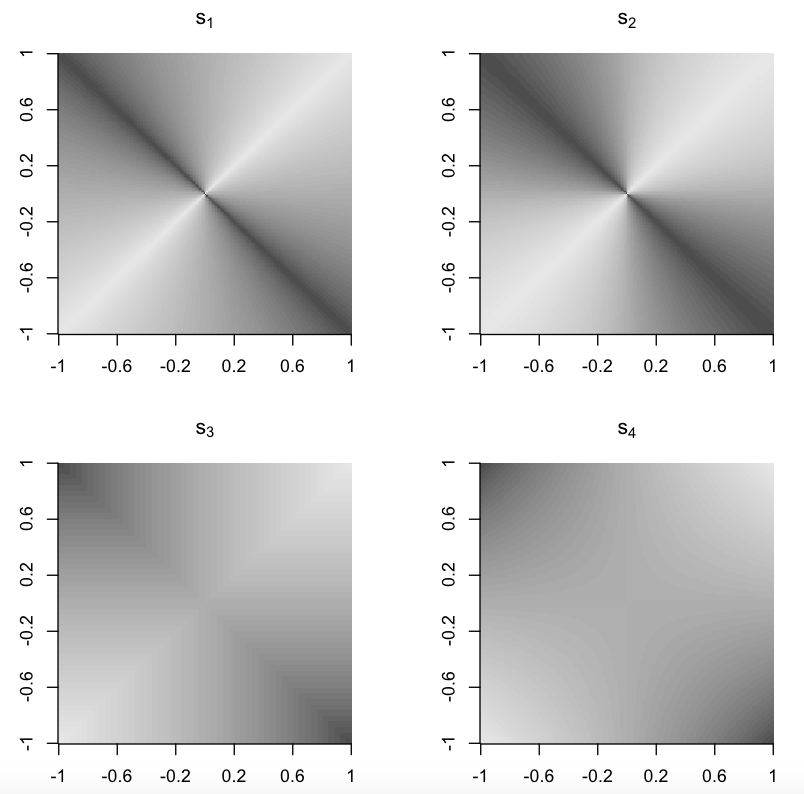}  
    \caption{The values of the four proposed similarity indices in the 
    region bound by $x \in [-1,1]$ and $y \in [-1,1]$.  The gray scale
    varies from 0 to 1. Observe the peak main diagonal in cases (a)
    and (b), which have a direct relationship with the Kronecker's delta
    and can be understood as a respectively smoothed version.}
    \label{fig:similarities}
    \end{center}
\end{figure}

Figure~\ref{fig:similarities} illustrates the four proposed similarity indices
in the region bound by $x \in [-1,1]$ and $y \in [-1,1]$.   Both $s_1$ (a)
and $s_2$ (b) yields marked peaks with value 1 along the main diagonal,
indicating the close relationship between these two similarity indices and
the Kronecker's delta function.  This diagonal peak is much less marked in 
the case of $s_3$ (c), and virtually undistinguishable in $s_4$.  

\vspace{0.5cm}

Perhaps the most important interpretation of Figure~\ref{fig:similarities}
consists in the fact that the similarity measurements $s_4$ --- upon
which the inner product and standard convolution, correlation, 
as well as the covariance statistical concept are based --- 
does not penalize the  region adjacent to the secondary 
diagonal $y=-x$.  It is precisely this characteristic of this index
that accounts for the tendency of the Pearson correlation coefficient
(e.g.~\cite{johnson:2002}),
which is derived from this index, to overestimate the joint variation
in cases where the two variables are not strongly related~\cite{CostaJaccard}.

\begin{figure}[h!]  
\begin{center}
   \includegraphics[width=1\linewidth]{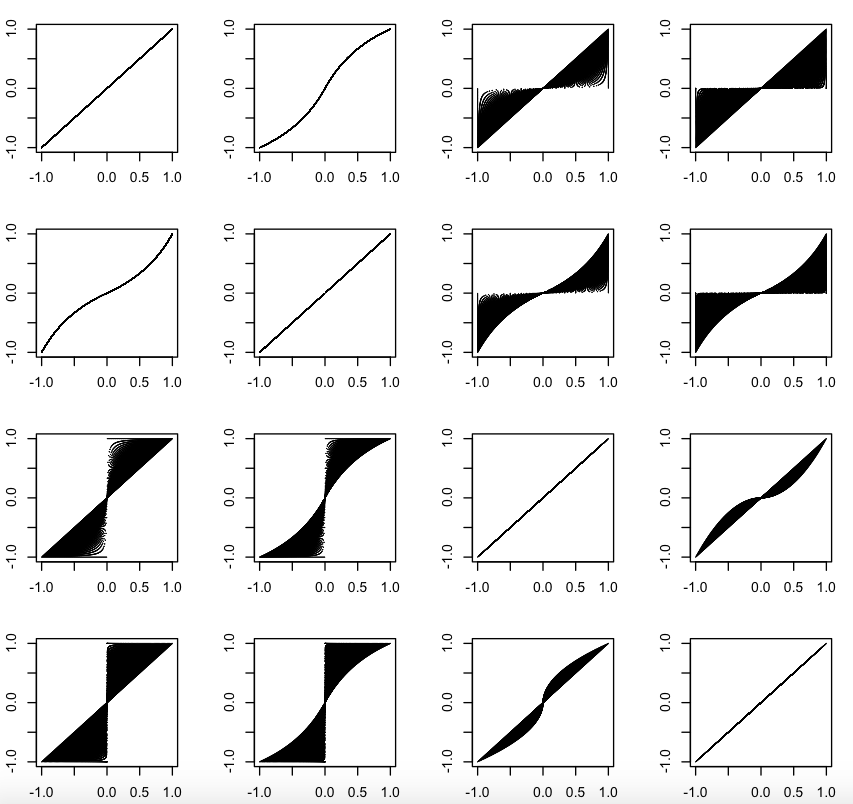}  
    \caption{Scatterplots illustrating the joint relationship between
    each pair of the four similarity indices $s_1$,
     $s_2$, $s_3$ and $s_4$.}
    \label{fig:correls}
    \end{center}
\end{figure}

Figure~\ref{fig:correls} illustrates scatterplots obtained with respect
to each pairwise association between the four indices $s_1$,
$s_2$, $s_3$ and $s_4$.  Interestingly, the first two indices are closely
related and bijectively associated.  This is not the case with the fourth
and fifth indices which, though similar (consider the respective scatterplot 
in the figure) are not bijective one another and much less with the other
two indices.

\vspace{0.5cm}

\section{Multiset Similarities}

Now that we have developed a principled approach to quantifying the
similarity between two real values $x$ and $y$, it becomes possible to
extend these indices to other mathematical structures, including 
multisets, vectors, functions, etc.  In this section we address the important
subject of quantifying the similarity between two multisets, which are henceforth
referred to as msets.

A multiset $A$ can be represented as:
\begin{equation}
   A = \lbar \left[ a_1, m_A(a_1) \right]; \left[ a_2, m_A(a_2) \right];  
     \ldots; \left[ a_N, m_A(a_N) \right] \rbar  \nonumber
\end{equation}

where we have $N$ elements $a_i$, each with respective multiplicity $m_A(a_i)$.
The support of this multiset is $S_A = \left\{ a_1, a_2, \ldots, a_N \right\}$.

The \emph{union} of two msets $A$ and $B$ sharing the same support is defined as:
\begin{eqnarray}
   A \cup B =  \lbar \left[ a_1, \max \left\{ m_A(a_1), m_B(a_1) \right\} \right];  \nonumber \\
   \left[ a_2, \max \left\{ m_A(a_2), m_B(a_2) \right\} \right];  \ldots;  \nonumber \\
   \left[ a_N, \max \left\{ m_A(a_N), m_B(a_N) \right\} \right] \rbar
\end{eqnarray}

In case $A$ and $B$ do not share the same support, a respective support can be
obtained for the mset union consisting of the set union of the respective mset supports.

The \emph{intersection} between two msets $A$ and $B$ sharing the same support is 
given as:
\begin{eqnarray}
   A \cap B =  \lbar \left[ a_1, \min \left\{ m_A(a_1), m_B(a_1) \right\} \right];  \nonumber \\
   \left[ a_2, \min \left\{ m_A(a_2), m_B(a_2) \right\} \right];  \ldots;  \nonumber \\
   \left[ a_N, \min \left\{ m_A(a_N), m_B(a_N) \right\} \right] \rbar
\end{eqnarray}

Msets can be generalized to real multiplicities, including possibly negative 
values~\cite{CostaJaccard,CostaMset}.

Quantification indices of the elementwise similarity between two multisets $A$ and $B$ can be
immediately obtained by applying the four scalar similarity indices proposed in
the previous section.

For simplicity's sake, we shall abbreviate $m_A(a_i)$ as $x_i$, and $m_B(a_i)$ as $y_i$,
which then yields:
\begin{align}
     &s_1(x_i,y_i) =  s_{x_i y_i} \frac{\min\left\{s_{x_i} x_i,  s_{y_i} y_i \right\} }{\max\left\{ s_{x_i} x_i, s_{y_i} y_i \right\} }   \\
     &s_2(x_i,y_i) = s_{x_i y_i}\frac{2 \min\left\{ s_{x_i} x_i,  s_{y_i} y_i \right\} }{s_{x_i} x_i + s_{y_i} y_i }    \\
     &s_3(x_i,y_i) =  \frac{  x_i  y_i }{\left( \max \left\{   x_i,    y_i \right\} \right)^2 }    \\
     &s_4(x_i,y_i) =  x_i y_i
\end{align}
 
It is of particular interest to generalize the four indices to quantify the similarity between
two multisets $A$ and $B$, which can be done as:
\begin{align}
     &s_1(A,B) =   \frac{ \sum_i s_{x_i y_i}\min\left\{s_{x_i} x_i,  s_{y_i} y_i \right\} }{\sum_i \max\left\{ s_{x_i} x_i, s_{y_i} y_i \right\} }   \\
     &s_2(A,B) = \frac{2 \sum_i s_{x_i y_i} \min\left\{ s_{x_i} x_i,  s_{y_i} y_i \right\} }
      {\sum_i  \left[ s_{x_i} x_i + s_{y_i} y_i \right] }    \\
     &s_3(A,B) = \frac{ \sum_i    x_i  y_i }
     {\sum_i \left( \max \left\{ s_{x_i}  x_i,   s_{y_i} y_i \right\} \right)^2 }    \\
     &s_4(A,B) =  \frac{ \sum_i x_i y_i } {|A| |B|}
\end{align}

The resulting index $s_1$ for msets is corresponds to the generalization of the
Jaccard similarity index to negative values~\cite{CostaJaccard,CostaMset} $J_N$,
while the index $s_2$ results identical to the also recently proposed 
addition-based mset Jaccard index (e.g.~\cite{CostaJaccard,CostaMset}), i.e.:
\begin{equation}
  s_1(A,B) = J_N(A,B)
\end{equation}

Therefore, the developments above allowed a principled derivation of those
recently introduced generalizations of the Jaccard similarity index
(e.g.~\cite{Jaccard, Jac:wiki}).

It also follows from the above developments that the generalized Jaccard
index can be understood as an implementation of smoothed generalizations
of  the Kronecker's delta function based similarity to scalars, msets, vectors
and scalar fields.

\section{Vector Similarities}

Since vectors can be understood as particular cases of msets with support 
$S = \lbar 1, 2, \ldots, N \rbar$, the respective generalization of the proposed
similarity to this type of mathematical structures is immediate.

Let two vectors $ \vec{x} = \left[ x_1, x_2, \ldots, x_N \right]$ and
$ \vec{y} = \left[y_1, y_2, \ldots, y_N \right]$.  We then have:
\begin{align}
     &s_1(x_i,y_i) =  s_{x_i y_i} \frac{\min\left\{s_{x_i} x_i,  s_{y_i} y_i \right\} }{\max\left\{ s_{x_i} x_i, s_{y_i} y_i \right\} }   \\
     &s_2(x_i,y_i) = s_{x_i y_i}\frac{2 \min\left\{ s_{x_i} x_i,  s_{y_i} y_i \right\} }{s_{x_i} x_i + s_{y_i} y_i }    \\
     &s_3(x_i,y_i) =  \frac{  x_i  y_i }{\left( \max \left\{   x_i,    y_i \right\} \right)^2 }    \\
     &s_4(x_i,y_i) =  x_i y_i
\end{align}

from which:
\begin{align}
     &s_1(\vec{x},\vec{y}) =   \frac{ \sum_i s_{x_i y_i}\min\left\{s_{x_i} x_i,  s_{y_i} y_i \right\} }{\sum_i \max\left\{ s_{x_i} x_i, s_{y_i} y_i \right\} }   \\
     &s_2(\vec{x},\vec{y}) = \frac{2 \sum_i s_{x_i y_i} \min\left\{ s_{x_i} x_i,  s_{y_i} y_i \right\} }
      {\sum_i  \left[ s_{x_i} x_i + s_{y_i} y_i \right] }    \\
     &s_3(\vec{x},\vec{y}) = \frac{ \sum_i  x_i  y_i }
     {\sum_i \left( \max \left\{ s_{x_i}  x_i,   s_{y_i} y_i \right\} \right)^2 }    \\
     &s_4(\vec{x},\vec{y}) =  \frac{ \sum_i x_i y_i } {|\vec{x}| |\vec{y}|}
\end{align}

Observe that the index $s_4$ becomes identical to the inner product between the two vectors.

\section{Function Similarities}

The generalization of the similarity indices to real functions follows directly from the
mset continuous representation~\cite{CostaJaccard,CostaMset}.

Given two real-valued functions $f(x)$ and $g(x)$ with shared support $S$, we 
immediately have:
\begin{align}
     &s_1(f(x),g(x)) =  s_{x_{f(x)g(x)}} \frac{\min\left\{s_{f(x)} f(x),  s_{g(x)} g(x) \right\} }
         {\max\left\{ s_{f(x)} f(x), s_{g(x)} g(x) \right\} }   \\
     &s_2(f(x),g(x)) = s_{f(x)g(x)}\frac{2 \min\left\{ s_{f(x)} f(x),  
          s_{g(x)} g(x) \right\} }{s_{f(x)} f(x) + s_{g(x)} g(x) }    \\
     &s_3(f(x),g(x))  =  \frac{  f(x) g(x) }{\left( \max \left\{   f(x),    g(x) \right\} \right)^2 }    \\
     &s_4(f(x),g(x))  =  f(x) g(x)
\end{align}

So that the respective functionals can be written as:
\begin{align}
     &s_1(f,g) =   \frac{ \int_S s_{f g} \min\left\{s_{f} f,  s_{g} g \right\} dx}{\int_S \max\left\{ s_{f} f, s_{g} g \right\} dx}   \label{eq:s1_func}  \\
     &s_2(f,g) = \frac{2  \int_Ss_{f g} \min\left\{ s_{f} f,  s_{g} g \right\} dx}
      {\int_S \left[ s_{f} f + s_{g} g \right] dx}    \\
     &s_3(f,g) = \frac{ \int_S  f  g \ dx}
     {\int_S \left( \max \left\{ s_{f}  f,   s_{g} g \right\} \right)^2 dx}    \\
     &s_4(f,g) =  \frac{ \int_S f g \ dx } {|f| |g|}
\end{align}

where $f(x)$ has been abbreviated as $f$, $g(x)$ has been abbreviated as $g$, 
 $|f| = \int_{S} |f(x)| dx$ and $|g| = \int_{S} |g(x)| dx$.

Another interesting implication of the similarity indices generalized to real
functions is that they can be understood as corresponding to binary
operations, in particular the product, between two functionals.

For instance, in the case of Equation~\ref{eq:s1_func}, we can write:
\begin{equation}
   s_1(f,g) =   \frac{ \ll f, g \gg }{ f \diamond g}    
\end{equation}

where:
\begin{equation}
    \ll f, g \gg  \ = \int_{S} s_{fg} \min \left\{ s_f f, s_g g \right \} dx
\end{equation}

which has been called the \emph{common product} between $f$ and 
$g$~\cite{CostaJaccard,CostaMset,CostaComparing,CostaAnalogies}.

and:
\begin{equation}
     f \diamond g  = \int_{S} \max \left\{ s_f f, s_g g \right \} dx
\end{equation}

which can be understood as a functional acting on the union of the
absolute valued versions of $f$ and $g$.

The other three similarities indices imply similar decompositions.

The functionals $\ll f, g, \gg$, $f \diamond g$ 
act on the following elementwise respective products:
\begin{eqnarray}
     \ll f, g \gg_p\ =  s_{fg} \min \left\{ s_f f, s_g g \right \} \\
     (f \diamond g)_p  =  \max \left\{ s_f f, s_g g \right \} 
\end{eqnarray}

and it is also possible to define the elementwise operations
related to the common product as:
\begin{equation}
     s_{1,p}(f,g) =    \frac{  s_{f g} \min\left\{s_{f} f,  s_{g} g \right\} }{ \max\left\{ s_{f} f, s_{g} g \right\} }   \
\end{equation}

However, observe that the functional $s_1(f,g)$ involves taking separated
integrals of the numerator and denominator.  

\begin{figure}[h!]  
\begin{center}
   \includegraphics[width=1\linewidth]{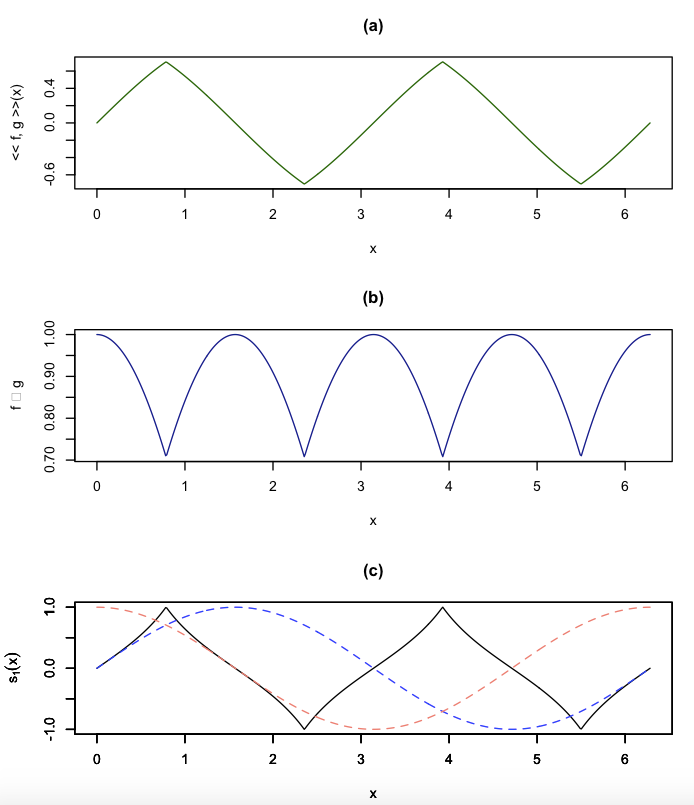}  
    \caption{The elementwise operations  $\ll f, g, \gg_p$, $f \diamond g_p$ and $s_{1,p}(f,g)$
    obtained for a complete period of the sine and cosine functions.}
    \label{fig:sincos}
    \end{center}
\end{figure}

Figure~\ref{fig:sincos} shows the elementwise operations
$\ll f, g \gg_p$,  $(f \diamond g)_p$, and $ s_{1,p}(f,g)$
in the case of the sine and cosine function within 
one complete period.  Observe these two functions shown as dashed
lines in Figure{fig:sincos}(c).  
We can observe that $\ll f, g\gg_p$ reveals to the common area between
the sine and cosine, while $(f diamond g)_p$ is the maximum between the
respective absolute values of these two functions.  The result
$s_1(f,g)$, provides an effective indication of the
signed similarity between the two considered functions.

\vspace{0.5cm}

\section{The Interiority and Coincidence Indices}

As shown in~\cite{CostaJaccard}, the traditional Jaccard similarity index between
two sets $A$ and $B$ is not capable of taking into account how much one of the sets
is interior to the other.  In order to compensate for this issue, a new similarity index,
called \emph{coincidence index} was proposed~\cite{CostaJaccard} as corresponding
to the product between the traditional Jaccard index and the interiority (or homogeneity)
index.

The interiority index can be expressed as:
\begin{equation}
   I(A,B) = \frac{A \cap B} {\min\left\{| A|, |B| \right\}}
\end{equation}

where $|A|$ and $|B|$ are the cardinalities of sets $A$ and $B$.

So that the coincidence index results as:
\begin{equation}
   C(A,B) = I(A,B) J(A,B) 
\end{equation}

Where $J(A,B)$ is the conventional Jaccard index.

Both the interiority and coincidence indices can also be understood as
corresponding to quantifications of similarities.  As such, it becomes interesting
to consider their generalizations to msets, vectors, and 
functions~\cite{CostaJaccard,CostaMset,CostaAnalogies}.

First, we consider the respective version of the interiority index allowing real multiplicities~\cite{CostaJaccard,CostaMset,CostaComparing,CostaAnalogies}:

In the case of msets, we have:
\begin{eqnarray}
   I(A,B) = \frac{\sum_{i \in S_+} \min\left\{s_{x_i} x_i, s_{y_i} y_i \right\}} 
   {\min\left\{\sum_{i \in S_+} s_{x_i} x_i , \; \sum_{i \in S_+} s_{y_i} y_i \right\}}  \\
\end{eqnarray}

where $S_+$  is the support restricted to the situations in which $s_{x_i} s_{y_i} >0$.
This restriction reflects the fact the understanding that it is impossible to have
interiority between two msets with all respective elements having opposite sign
multiplicities.  Observe that, as a consequence, $0 \leq I(A,B) \leq 1$.

In case the whole support is to be taken into account, which can be required
in some circumstances such as when performing template
matching~\cite{CostaComparing}, we can make:
\begin{eqnarray}
   I(A,B) = \frac{\sum_{i \in S} \min\left\{s_{x_i} x_i, s_{y_i} y_i \right\}} 
   {\min\left\{\sum_{i \in S} s_{x_i} x_i , \; \sum_{i \in S} s_{y_i} y_i \right\}}  \\
\end{eqnarray}

In the case of vectors, we immediately have:
\begin{eqnarray}
   I(\vec{x},\vec{y}) = \frac{\sum_{i \in S_+} \min\left\{s_{x_i} x_i, s_{y_i} y_i \right\}} 
   {\min\left\{\sum_{i \in S_+} s_{x_i} x_i , \; \sum_{i \in S_+} s_{y_i} y_i \right\}}  \\
\end{eqnarray}

So that:
\begin{equation}
   C(\vec{x}, \vec{y}) =  I(\vec{x},\vec{y}) J(\vec{x},B\vec{y}) = I(\vec{x},\vec{y}) \; s_1(\vec{x},B\vec{y})
\end{equation}

And, for functions:
\begin{eqnarray}
   I(f,g) = \frac{\int_{S_+} \min\left\{s_{f} f, s_{g} g \right\} dx} 
   {\min\left\{\int_{S_+} s_{f} f dx , \; \int_{S_+} s_{g} g dx \right\}}  \\
\end{eqnarray}

Implying:
\begin{equation}
   C(f, g) =  I(f,g) J(f,g) = I(f,g) \; s_1(f,g)
\end{equation}

\section{Similarity Convolutions and Correlations}

Each of the similarity indices generalized to the real space of functions corresponds
to a valid functional.  Now, it is possible to obtain respective \emph{convolutions}
and \emph{correlations}.  For instance, in the case of $s_1$, we have the
following respectively associated convolution:
\begin{equation}
    (f \Box g)_{s1}[y] =   \frac{ \int_S s_{f g} \min\left\{s_{f} f,  s_{g} g(y-x) \right\} dx}
    {\int_S \max\left\{ s_{f} f, s_{g} g(y-x) \right\} dx}   \\
\end{equation}

and correlation:
\begin{equation}
    (f \blacksquare g)_{s1}[y] =   \frac{ \int_S s_{f g} \min\left\{s_{f} f,  s_{g} g(x-y) \right\} dx}
    {\int_S \max\left\{ s_{f} f, s_{g} g(x-y) \right\} dx}   \\
\end{equation}

In the case of $s_4$, we have:
\begin{equation}
    (f \Box g)_{s4}[y] =   \frac{ \int_S   s_{g} g(y-x)  dx}
    { |f| |g|  }   \\
\end{equation}

which corresponds to a normalized version of the standard convolution.

Since the common product
is associated to the Walsh functions~\cite{CostaAnalogies}, it
is possible to perform convolutions involving these functions
(e.g.~\cite{Walsh, Stoffer, Tzafestas, Harmuth})
by using fast  computing schemes (e.g.~\cite{FastWalsh}) analogous to the fast 
Fourier transform (e.g.~\cite{brigham:1988}). 

Another interesting point is that, while the products in the Fourier transform
are bilinear, the analogous counterpart in the common product is the
non-linear operation of maximum.  This confers some important properties
to respective related operations, such as correlation, such as substantially
enhanced performance in tasks such as template matching and 
filtering~\cite{CostaComparing}.

\section{Concluding Remarks}

The concept of similarity appears recurrently in science and technology, underlying
a large number of concepts, operations, and properties.  From the perspective of
Hilbert spaces, the similarity is critically important as it is related to the concept
of inner product on which those spaces are based.  However, the quantification of
similarities between mathematical entities also constitute an ubiquitous task in
virtually every applied area, including but by no means limited to patter recognition,
signal processing, and machine intelligence, to name but a few cases.

In the present work, we developed a principled approach in which the Kronecker's
delta function was taken as the prototypical reference for quantifying the similarity
between two scalar values, and then developed more yielding versions involving
the operations of minimum, maximum, sum and product, in addition to the sign
function.   Four main indices were obtained, three of which are normalized in
the interval $[0,1]$, which were then extended to respective signed versions
capable of providing more information about the kind of similarity, yielding
respective versions of these indices bound by the interval $[-1,1]$. 

Then, relying on recent results regarding the extension of multisets to
functions and other mathematical structures~\cite{CostaJaccard,CostaMset,CostaAnalogies},
we were able to extend the four signed similarity indices to multisets, vectors,
and then functions.  The extension to other mathematical structures including scalar
and vector fields can also be obtained in analogous manner.

Several important results have been obtained.  First, we have that the extensively
applied Jaccard index relates directly to the similarity index $s_1$, while the
index $s_4$ let to the standard inner product functional and convolution.
Of particular interest is that the similarity functionals recently introduced 
in~\cite{CostaJaccard,CostaMset,CostaComparing,CostaCCA} resulted 
naturally from the here reported developments.  For instance, it has been
possible to verify that the mset Jaccard index, when adapted to negative
values, corresponds to the functional respective to the described index $s_1$.  
In addition, the addition-based mset Jaccard index was shown to follow
from the index $s_2$.  The index $s_4$, which is unbound, led to the
standard inner product and respectively associated convolution and correlation.

\vspace{0.7cm}
\emph{Acknowledgments.}

Luciano da F. Costa
thanks CNPq (grant no.~307085/2018-0) and FAPESP (grant 15/22308-2).  
\vspace{1cm}

\bibliography{mybib}
\bibliographystyle{unsrt}

\end{document}